\documentclass[10pt,twocolumn,letterpaper]{article}

\usepackage{tango}
\usepackage{float}

\usepackage[pagenumbers]{cvpr}

\usepackage[numbers]{natbib}
\usepackage{graphicx}
\usepackage{amsmath}
\usepackage{amssymb}
\usepackage{booktabs}
\usepackage{tabularx}
\usepackage[pagebackref,breaklinks,colorlinks]{hyperref}
\usepackage[capitalize]{cleveref}
\usepackage{textcomp}

\usepackage{wrapfig}

\makeatletter
\@namedef{ver@everyshi.sty}{}
\makeatother
\usepackage{tikz}
\usepackage{pgfplots}

\crefname{section}{Sec.}{Secs.}
\Crefname{section}{Section}{Sections}
\Crefname{table}{Table}{Tables}
\crefname{table}{Tab.}{Tabs.}

\def\cvprPaperID{9157}
\def\confName{CVPR}
\def\confYear{2022}

\begin{document}

\title{Cross-view Transformers for real-time Map-view Semantic Segmentation}

\author{
    Brady Zhou\\
    UT Austin\\
    {\tt\small brady.zhou@utexas.edu}
    \and
    Philipp Kr{\"a}henb{\"u}hl \\
    UT Austin \\
    {\tt\small philkr@cs.utexas.edu}
}
\maketitle

\begin{abstract}
  We present cross-view transformers, an efficient attention-based model for map-view semantic segmentation from multiple cameras.
  Our architecture implicitly learns a mapping from individual camera views into a canonical map-view representation using a camera-aware cross-view attention mechanism.
  Each camera uses positional embeddings that depend on its intrinsic and extrinsic calibration.
  These embeddings allow a transformer to learn the mapping across different views without ever explicitly modeling it geometrically.
  The architecture consists of a convolutional image encoder for each view and cross-view transformer layers to infer a map-view semantic segmentation.
  Our model is simple, easily parallelizable, and runs in real-time.
  The presented architecture performs at state-of-the-art on the nuScenes dataset, with 4x faster inference speeds.
  Code is available at \url{https://github.com/bradyz/cross_view_transformers}.
\end{abstract}

\section{Introduction}

Autonomous vehicles depend on robust scene understanding and online mapping to navigate the world.
To drive safely, these systems not only reason about the semantics of their surroundings, but also a spatial understanding due to the geometric nature of navigation.
Many prior approaches directly model geometry and relationships between different view and a canonical map representation~\cite{longuet1981computer,kanade1994stereo,schoenberger2016sfm,ammar2019geometric,kim2019deep,philion20,fiery2021}.
They require an explicit~\cite{longuet1981computer,kanade1994stereo,schoenberger2016sfm,ammar2019geometric,kim2019deep} or probabilistic~\cite{philion20,fiery2021} estimate of depth in either image or map-view.
However, this explicit modeling can be hard.
First, image-based depth estimates are error-prone, as monocular depth estimates scale poorly with the distance to the observer.
Second, depth-based projections are a fairly inflexible and rigid bottleneck to map between views.
In this work, we take a different approach.

We learn to map from camera-view to a canonical map-view representation using a cross-view transformer architecture.
The transformer does not perform any explicit geometric reasoning but instead learns to map between views through a geometry-aware positional embedding.
Multi-head attention then learns to map features from camera-view into a canonical map-view representation using a learned map-view positional embedding.
We learn a single map-view positional embedding for all cameras and perform attention across all views.
The model thus learns to link up different map locations to both cameras and locations within each camera.
Our cross-view transformer refines the map-view embedding through multiple attention and MLP blocks.
The cross-view transformer allows the network to learn any geometric transformation implicitly and directly from data.
It learns an implicit estimate of depth through the camera-dependent map-view positional embedding by performing the downstream task as accurately as possible.

\begin{figure}[t]
\centering
\includegraphics[page=1,width=\linewidth]{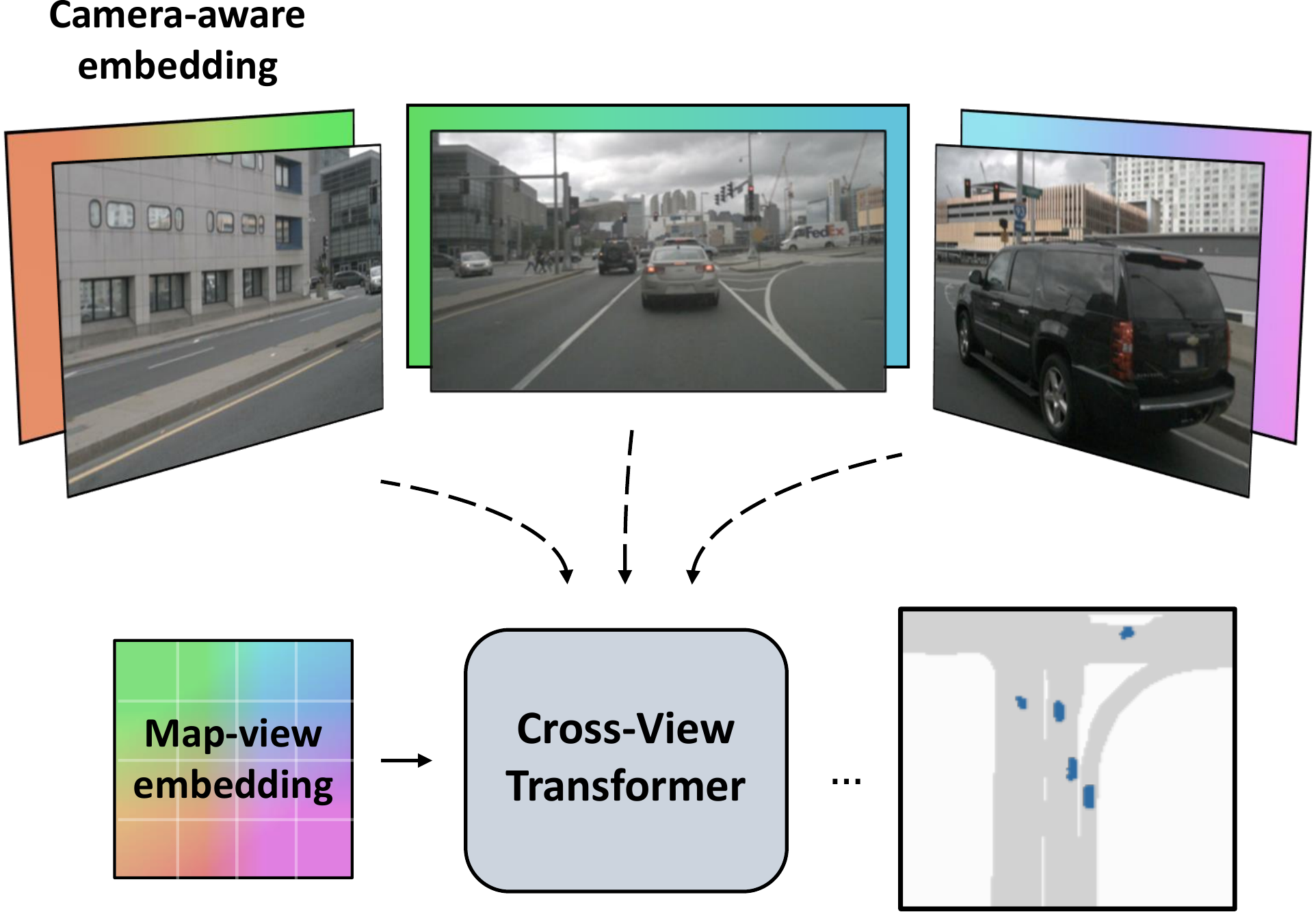}%
\caption{We introduce an architecture for perception in a map-view frame from multiple views. Our model builds a map-view representation by cross-attending to image features. A camera-aware positional embedding can geometrically link up the camera and map-views.}
\end{figure}

The simplicity of our model is a key strength.
The model performs at state-of-the-art on the nuScenes~\cite{nuscenes2019} dataset for vehicle and road segmentation in the map-view and comfortably runs in real-time (35 FPS) on a single RTX 2080 Ti GPU.
The model is easy to implement and trains within 32 GPU hours.
The learned attention mechanism learns accurate correspondences between camera and map-view directly from data.

\section{Related Works}

Map-view semantic segmentation lies at the intersection of 3D recognition, depth estimation, and mapping.

\paragraph{Monocular 3D object detection.}

Monocular detection aims to find objects in a scene, estimate their real-world size, orientation, and placement in the 3D scene.
Most common approaches reduce the problem to 2D object detection and infer monocular depth~\cite{zhou2019objects,manhardt2019roi}.
CenterNet~\cite{zhou2019objects} directly predicts depth for each image coordinate.
ROI-10D~\cite{manhardt2019roi} lifts 2D detections into 3D using depth estimates then regresses 3D bounding boxes.
Psuedo-lidar based approaches~\cite{weng2019monocular, ma2019accurate, xu2018multi, wang2019pseudo} project to 3D points using a depth estimate and leverage 3D point based architectures (e.g.~\cite{qi2017pointnet, lang2019pointpillars, vora2020pointpainting}) with 2D labels.
This family of algorithms directly benefits from advances in monocular depth estimation and 3D vision.

Monocular 3D object detection is both easier and harder than mapping from multiple cameras.
The overall problem setup deals with just a single camera and does not need to merge multiple sources of inputs.
However, it strongly relies on a good explicit monocular depth estimate, which may be harder to obtain.

\paragraph{Depth estimation.}
Depth is a core ingredient in many multi-view mapping approaches.
Classic structure-from-motion approaches~\cite{longuet1981computer,agarwal2011building,snavely2006photo,frahm2010building,schoenberger2016sfm} leverage epipolar geometry and triangulation to explicitly compute camera extrinsics and depth.
Stereo matching finds corresponding pixels, from which depth can be explicitly computed~\cite{kanade1994stereo}.
Recent deep learning approaches directly regress depth from images~\cite{eigen2014depth, godard2017unsupervised, godard2019digging, zhou2017unsupervised, Ranftl2020, fu2018deep}.

While convenient, explicit depth is challenging to utilize for downstream tasks.
It is camera-dependent and requires an accurate calibration and fusion of multiple noisy estimates.
Our approach side-steps explicit depth estimation and instead allows an attention mechanism with positional embedding to take its place.
Our cross-view transformer learns to reproject camera views into a common map representation as part of training.

\paragraph{Semantic mapping in the map-view.}
Driven by ever larger 3D recognition datasets~\cite{KITTI,cordts2015cityscapes,nuscenes2019,sun2020waymo,houston2021one}, a number of works have focused on perception in the map-view.
This problem is particularly challenging as the inputs and outputs lie in different coordinate frames.
Inputs are recorded in calibrated camera views, outputs are rasterized onto a map.
Most prior works differ in the way the transformation is modeled.
One common technique is to assume the scene is mostly planar and represent the image to map-view transformation as a simple homography~\cite{garnett20193d, loukkal2021driving, kim2019deep, zhu2021monocular, sengupta2012automatic, ammar2019geometric}.
A second family of methods directly produces map-view predictions from input images, with no explicit geometric modeling.
VED~\cite{lu2019monocular} uses a Variational Auto Encoder~\cite{kingma2013auto} to produce a semantic occupancy grid from a single monocular camera-view.

Closely related in spirit to our method, VPN~\cite{pan2020cross} learns a common feature representation across multiple views with their proposed \textit{view relation module} - an MLP that outputs map-view features from inputs across all views.
Both VED and VPN show carefully-designed networks trained with sufficient training data can jointly learn the map-view transformation and perform prediction.
However, these methods do suffer certain drawbacks as they do not model the geometric structure of the scene.
They forgo the inherit inductive biases contained in a calibrated camera setup and instead need to learn an implicit model of camera calibration baked into the network weights.
Our cross-view transformer instead uses positional embeddings derived from calibrated camera intrinsics and extrinsics.
The transformer can learn a camera-calibration-dependent mapping akin to raw geometric transformations.

Most recently, top-performing methods returned back to explicit geometric reasoning~\cite{roddick2018orthographic, philion20, roddick2020predicting, saha2021enabling, murez2020atlas, fiery2021}.
Orthographic Feature Transform (OFT)~\cite{roddick2018orthographic} creates a map-view intermediate representation from a monocular image by average pooling image features from the 2D projection that corresponds with the pillar in map-view.
This pooling operation foregoes an explicit depth estimate and instead averages all possible image locations a map-view object could take.
Lift-Splat-Shoot (LSS)~\cite{philion20} constructs an intermediate map-view representation in a similar fashion.
However, they allow the model to learn a soft depth estimate and average across different bins using a learned depth-estimate-dependent weight.
Their downstream decoder can account for uncertainty in depth.
This weighted averaging operation closely mimics the attention used in a transformer.
However, their ``attention weights'' are derived from geometric principles and not learned from data.
The original Lift-Splat-Shoot approach considers multiple views within a single timestep.
Recent methods have extended this further to take aggregate features from previous timesteps~\cite{saha2021enabling}, and use multi-view, multi-timestep observations to do motion forecasting~\cite{fiery2021}.

In this work, we show that implicit geometric reasoning performs as well as explicit geometric models.
The added benefit of our implicit handling of geometry is an improvement in inference speed compared to explicit models.
We simply learn a set of positional embeddings, and attention will reproject the camera to map-view.

\begin{figure*}[t]
\centering
\includegraphics[page=1,width=0.85\linewidth]{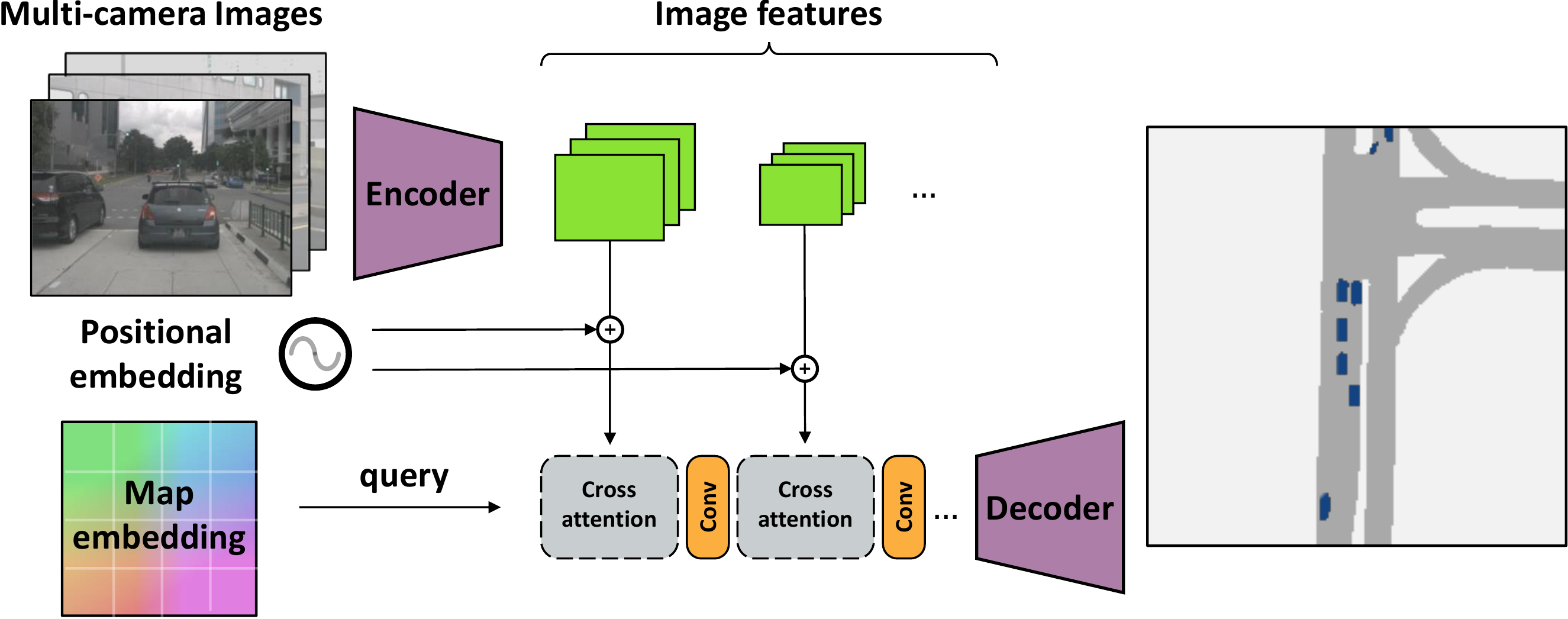}
\caption{An overview of our proposed architecture for map-view segmentation. For each image, we extract image features across multiple scales. Using known camera pose and intrinsics, we construct a camera-aware positional embedding. We learn a map-view positional embedding that aggregates information from all views through a series of cross attention layers.
Each cross-attention refines the map-view positional embedding and queries more accurate image locations.}
\label{fig:method}
\end{figure*}

\section{Cross-view transformers}
In this section, we introduce our proposed architecture for semantic segmentation in the map-view from multiple camera views.
In this task, we are given a set of $n$ monocular views ${(I_k, K_k, R_k, t_k)}_{k=1}^n$ consisting of input image $I_k \in \mathbb{R}^{H \times W \times 3}$, camera intrinsics $K_k \in \mathbb{R}^{3 \times 3}$, and extrinsic rotation $R_k \in \mathbb{R}^{3 \times 3}$ and translation $t_k \in \mathbb{R}^{3}$ relative to the center of the ego-vehicle.
Our goal is to learn an efficient model to extract information from the multiple camera views in order to predict a binary semantic segmentation mask $y \in {\{0, 1\}}^{h \times w \times C}$ in the orthographic map-view coordinate frame.

We design a simple, yet effective encoder-decoder architecture for map-view semantic segmentation.
An image-encoder produces a multi-scale feature representation of each input image.
A cross-view cross-attention mechanism then aggregates multi-scale features into a shared map-view representation.
The cross-view attention relies on a positional embedding that is aware of the geometric structure of the scene and learns to match up camera-view and map-view locations. 
All cameras share the same image-encoder, but use a positional embedding dependent on their individual camera calibration.
Finally, a lightweight convolutional decoder upsamples the refined map-view embedding and produces the final segmentation output.
The entire network is end-to-end differentiable and learned jointly.
Figure~\ref{fig:method} shows an overview of the full architecture.

In Section~\ref{sec:mvt}, we first present the core cross-view attention mechanism and positional embedding that underlies our entire architecture.
Section~\ref{sec:arch} then combines multiple cross-view attention layers into the final map-view segmentation model.

\subsection{Cross-view attention}
\label{sec:mvt}
The goal of cross-view attention is to link up a map-view representation with image-view features.
For any world coordinate $x^{(W)} \in \mathbb{R}^3$, the perspective transformation describes its corresponding image coordinate $x^{(I)} \in \mathbb{R}^3$:
\begin{equation}
 x^{(I)} \simeq K_k R_k (x^{(W)} - t_k) \label{eq:perspective}.
\end{equation}
Here, $\simeq$ describes equality up to a scale factor, and $x^{(I)}=(\cdot,\cdot,1)$ uses homogeneous coordinates.
However, without an accurate depth estimate in camera view or height-above-ground estimate in map-view, the world coordinate $x^{(W)}$ is ambiguous.
We do not learn an explicit estimate of depth but encode any depth ambiguity in the positional embeddings and let a transformer learn a proxy for depth.

We start by rephrasing the geometric relationship between world and image coordinates in Equation~\ref{eq:perspective} as a cosine similarity for use in an attention mechanism.
\begin{equation}
 sim_k(x^{(I)}, x^{(W)}) = \frac{\left(R_k^{-1} K_k^{-1} x^{(I)}\right) \cdot \left(x^{(W)} - t_k\right)}{\|R_k^{-1} K_k^{-1} x^{(I)})\|\|(x^{(W)} - t_k\|} \label{eq:similarity}.
\end{equation}
This similarity still relies on the exact world coordinate $w^{(W)}$.
Next, we replace all geometric components of this similarity with positional encodings that can learn both geometric and appearance features.

\paragraph{Camera-aware positional encoding.}
The camera-aware positional encoding starts from the unprojected image coordinate $d_{k,i} = R_k^{-1} K_k^{-1} x^{(I)}_i$ for each image coordinate $x^{(I)}_i$.
The unprojected image coordinate $d_{k,i}$ describes a direction vector from the origin $t_k$ of camera $k$ to the image plane at depth $1$.
The direction vector uses world coordinates.

We encode this direction vector $d_{k,i}$ using an MLP (shared across all $k$ views) into a $D$-dimensional positional embedding $\delta_{k,i} \in \mathbb{R}^{D}$.
We use $D=128$ in our experiments.
We combine this positional embedding with image features $\phi_{k,i}$ in the keys of our cross-view attention mechanism.
This allows cross-view attention to use both appearance and geometric cues to reason about correspondences across the different views.

Next, we show how to build an equivalent representation for the map-view queries.
This embedding can no longer rely on exact geometric inputs and instead needs to learn geometric reasoning in consecutive layers of the transformer.

\paragraph{Map-view latent embedding.}
The map-view component of the geometric similarity metric in Equation~\ref{eq:similarity} contains a world coordinate $x^{(W)}$ and camera location $t_k$.
We encode both in a separate positional embedding.
We use an MLP to transform each camera location $t_k$ into an embedding $\tau_k \in \mathbb{R}^D$.
We build the map-view representation up over multiple iterations in our transformer.
We start with a learned positional encoding $c^{(0)} \in \mathbb{R}^{w \times h \times D}$.
The goal of the map-view positional encoding is to produce an estimate of the 3D location of each element of the road.
Initially, this estimate is shared across all scenes and likely learns an average position and height above the ground plane for each element of the scene.
The transformer architecture then refines this estimate through multiple rounds of computation, resulting in new latent embeddings $c^{(1)}, c^{(2)}, \ldots$.
Each positional embedding is better able to project the map-view coordinates into a proxy of the 3D environment.
Following the geometric similarity measure in Equation~\ref{eq:similarity}, we use the difference between map-view embeddings $c$ and camera-location embeddings $\tau_k$ as queries in the transformer.

\paragraph{Cross-view attention.}
Our cross-view transformer combines both positional encodings through a cross-view attention mechanism.
We allow each map-view coordinate to attend one or more image locations.
Crucially, not every map-view location has a corresponding image patch in each view.
Front-facing cameras do not see the back, rear-facing cameras do not see the front.
We allow the attention mechanism to select both camera and location within each camera when corresponding map-view and camera-view perspectives.
To this end, we first combine all camera-aware positional embeddings $\delta_1, \delta_2, \ldots$ from all views into a single key vector $\delta = \left[\delta_1, \delta_2, \ldots \right]$.
At the same time, we combine all image features $\phi_1, \phi_2, \ldots$ into a single value vector $\phi = \left[\phi_1, \phi_2, \ldots\right]$.
We combine camera-aware positional embeddings $\delta$ and image features $\phi$ to compute attention keys.
Finally, we perform softmax-cross-attention~\cite{vaswani2017attention} between keys $[\delta, \phi]$, values $\phi$, and map-view queries $c - \tau_k$.

The softmax attention uses a cosine similarity between keys and queries as a basic building block
\begin{equation}
sim(\delta_{k,i}, \phi_{k,i}, c_j^{(n)}\!, \tau_k) = \frac{\left(\delta_{k,i}+\phi_{k,i}\right)\!\cdot\!\left(c_j^{(n)}\!-\tau_k\right)}{\|\delta_{k,i}+\phi_{k,i}\|\|c_j^{(n)}\!-\tau_k\|}.
\end{equation}
This cosine similarity follows the geometric interpretation in Equation~\ref{eq:similarity}.
This cross-view attention forms the basic building block of our cross-view transformer architecture.

\subsection{A cross-view transformer architecture}
\label{sec:arch}

The first stage of the network builds up a camera-view representation for each input image.
We feed each image $I_i$ into feature extractor (EfficientNet-B4~\cite{tan2019efficientnet}) and get a multi-resolution patch embedding $\{\phi_1^1, \phi_1^2, \ldots, \phi_n^R\}$, where $R$ is the number of resolutions we consider.
We found $R=2$ resolutions to produce sufficiently accurate results.
We process each resolution separately.
We start from the lowest resolution and project all image features into map-view using cross-view attention.
We then refine the map-view embedding and repeat the process for higher resolutions.
Finally, we use three up-convolutional layers to produce the full resolution output.

A detailed overview of this architecture is shown in Figure~\ref{fig:method}.
The final network is end-to-end trainable.
We train all layers using ground truth semantic map-view annotations and a focal loss~\cite{lin2017focal}.

\section{Implementation Details}

\paragraph{Architecture.}
We use (and fine-tune) a pre-trained EfficientNet-B4~\cite{tan2019efficientnet} to compute image features at two different scales - (28, 60) and (14, 30), which correspond to a 8x and 16x downscaling, respectively.
The initial map-view positional embedding is a tensor of learned parameters $w \times h \times D$, where $D = 128$.
For computational efficiency, we choose $w = h = 25$ as the cross-attention function grows quadratically with grid size.
The encoder consists of two cross-attention blocks: one for each scale of patch features.
We use multi-head attention with $4$ heads and an embedding size $d_{head} = 64$.
The decoder consists of three (bilinear upsample + conv) layers to upsample the latent representation to the final output size.
Each upsampling layer increases the resolution by a factor of 2 up to a final output resolution of $200 \times 200$.
This corresponds to a $100 \times 100$ meter area centered around the ego-vehicle.

\paragraph{Training.}
We train all models using a focal loss~\cite{lin2017focal}, with a batch size of 4 (per GPU) for 30 epochs.
We optimize using the AdamW~\cite{loshchilov2017decoupled} optimizer with the one-cycle learning rate scheduler~\cite{smith2019super}.
Training converges within 8 hours on a 4 GPU machine.

\section{Results}
\label{sec:results}

\begin{table}[h]
\centering
\begin{tabular}{l@{}c@{\ \ \ }c@{\ \ \ }c@{\ \ \ }c}
\toprule
 & Setting 1 & Setting 2 & \#Params (M) & FPS \\
\midrule
PON~\cite{roddick2020predicting}    & 24.7 & - & 38 & 30 \\
VPN~\cite{pan2020cross}             & 25.5 & - & 18 & - \\
STA~\cite{saha2021enabling}         & 36.0 & - & - & - \\
Lift-Splat~\cite{philion20}         & - & 32.1 & 14 & 25 \\
FIERY~\cite{fiery2021}              & \textbf{37.7} & \textbf{35.8} & 7 & 8 \\
Ours                                & \textbf{37.5} & \textbf{36.0} & \textbf{5} & \textbf{35} \\
\bottomrule
\end{tabular}
\caption{Vehicle map-view segmentation on nuScenes.
Setting 1 refers to the 100m$\times$50m at 25cm resolution setting proposed by Roddick \etal~\cite{roddick2020predicting}.
Setting 2 refers to the 100m$\times$100m at 50cm resolution setting proposed by Philion and Fidler~\cite{philion20}.
Both settings evaluate the Intersection over Union (IoU) metric.
Higher is better.
For a fair comparison, we use single-timestep models only. In particular, we compare to FIERY static~\cite{fiery2021}.
In both settings, our cross-view transformer performs at the state-of-the-art with a smaller model and runs $4.5\times$ faster during inference.}
\label{table:results}
\end{table}

We evaluate our cross-view transformer on vehicle and road map-view semantic segmentation on the nuScenes~\cite{nuscenes2019} and Argoverse~\cite{chang2019argoverse} datasets.

\paragraph{Dataset.}

The nuScenes~\cite{nuscenes2019} dataset is a collection of 1000 diverse scenes collected over a variety of weathers, time-of-day, and traffic conditions.
Each scene lasts 20 seconds and contains 40 frames for a total of 40k total samples in the dataset.
The recorded data captures a full 360\textdegree\ view around the ego-vehicle and is composed of 6 camera views.
Each camera view has calibrated intrinsics $K$ and extrinsics $(R, t)$ at every timestep.
We resize every image to $224 \times 448$ unless specified otherwise.
The Argoverse~\cite{chang2019argoverse} dataset contains 10k total frames.

Vehicles and other objects in the scene are tracked across frames and annotated with 3D bounding boxes using LiDAR data.
Using the pose of the ego-vehicle, we generate the ground-truth labels $y$, a binary vehicle occupancy mask rendered at a resolution of (200, 200) by orthographically projecting 3D box annotations onto the ground plane, following standard practice~\cite{fiery2021,philion20}.

\paragraph{Evaluation.}
There are two commonly used evaluation settings for map-view vehicle segmentation.
Setting 1 uses a 100m$\times$50m area around the vehicle and samples a map at a 25cm resolution.
The setting, popularized by Roddick \etal~\cite{roddick2020predicting}, serves as the main comparison to prior work.
Setting 2~\cite{philion20} uses a 100m$\times$100m area around the vehicle, with a 50cm sampling resolution.
This setting was popularized by Philion and Fidler~\cite{philion20} and serves as a comparison to Lift-Splat-Shoot~\cite{philion20} and FIERY~\cite{fiery2021}.
We use Setting 2 for all ablations.
In both settings, we use the Intersection-over-Union (IoU) score between the model predictions and the ground truth map-view labels as the main performance measure.
We additionally report inference speeds measured on an RTX 2080 Ti GPU.

\subsection{Comparison to prior work}

\begin{table}[t]
\centering
\begin{tabular}{l c c}
\toprule
 & Vehicle & Driveable Area \\
\midrule
OFT~\cite{roddick2018orthographic}    & 30.1 & 71.7 \\
Lift-Splat~\cite{philion20}           & 32.1 & 72.9 \\
Ours                                  & \textbf{36.0} & \textbf{74.3} \\
\midrule
Monolayout~\cite{mani2020monolayout}   & 32.1 & 58.3 \\
PON~\cite{roddick2020predicting}       & 31.4 & 65.4 \\
Ours                                   & \textbf{35.2} & \textbf{73.6} \\
\bottomrule
\end{tabular}
\caption{Additional comparison with models that perform map-view segmentation for vehicles and driveable area. The top and bottom rows correspond to on nuScenes~\cite{nuscenes2019} setting 2 and Argoverse~\cite{chang2019argoverse} dataset respectively.}
\end{table}

\begin{table}[b]
\centering
\begin{tabular}{l c}
\midrule
& IoU \\
\midrule
No camera-aware embedding $\delta$ & 31.0 \\
No image features $\phi$ in attention & 33.2 \\
No map-view embedding refinement & 33.6 \\
\midrule
Full model & 36.0 \\
\bottomrule
\end{tabular}
\caption{Ablations of the cross-view attention mechanism. The first row compares to a model that does not use camera-aware positional embedding thus only uses image features as attention keys.
The second row does not use any image features in the keys of the attention mechanism.
The third row uses the full attention computation in camera-view but does not refine the map-view positional embedding.
All partial models degrade reasonably and perform below the full model.}
\label{table:ablation}
\end{table}

We compare our model to the five most competitive prior approaches on online mapping.
For a fair comparison, we use single-timestep models only and do not consider temporal models.
We compare to Pyramid Occupancy Networks (PON)~\cite{roddick2020predicting}, Orthographic Feature Transform (OFT)~\cite{roddick2018orthographic}, View Parsing Network (VPN)~\cite{pan2020cross}, Spatio-temporal Aggregation (STA)~\cite{saha2021enabling}, Lift-Splat-Shoot~\cite{philion20}, and FIERY~\cite{fiery2021}.
PON, VPN, STA only report numbers in Setting 1, while Lift-Splat-Shoot only uses Setting 2.

In both settings, our cross-view transformer and FIERY outperform all alternative approaches by a significant margin.
Our cross-view transformer and FIERY perform comparably.
We have a slight edge in Setting 2, FIERY in Setting 1.
The main advantage of our model is simplicity and inference speed, along with the accompanying edge in model size.
Our model trains significantly faster (32 GPU hours vs 96 GPU hours) and performs $4\times$ faster inference.
We intentionally use the same image feature extractor (EfficientNet-B4)~\cite{tan2019efficientnet} and similar decoder architecture as FIERY.
This suggests our cross-view transformer is capable of combining features from multiple views in a more efficient manner.

\subsection{Ablations of cross-view attention}

The core ingredient of our approach is the cross-view attention mechanism.
It combines camera-aware embeddings and image features as keys and learned map-view positional embeddings as queries.
The map-view embeddings are allowed to update across multiple iterations, while the camera-aware embeddings contain some geometric information.
Table~\ref{table:ablation} compares the impact of each of the components of the attention mechanism on the resulting map-view segmentation system.
For each ablation, we trained a model from scratch using equivalent experimental settings, changing a single component at the time.

The most important component of our system is the camera-aware positional embedding.
It bestows the attention mechanism with the ability to reason about the geometric layout of the scene.
Without it, attention has to rely on the image feature to reveal its own location.
It is possible for the network to learn this localization due to the size of the receptive field and zero padding around the boundary of the image.
However, an image feature alone struggles to properly link up map-view and camera-view perspectives.
It also needs to explicitly infer the direction each image is facing to disambiguate different views.
On the other hand, a purely geometric camera-aware positional embedding alone is also insufficient.
The network likely uses both semantic and geometric cues to align map-view and camera-view, especially after the refinement of the map-view embedding.
Finally, using a single fixed map-view embedding also degrades the performance of the model.
The final model performs best with all its attention components.

\begin{table}[b]
\centering
\begin{tabular}{l c}
\midrule
Method & IoU \\
\midrule
None & 31.0 \\
Learned per camera & 34.4 \\
Camera-aware + Random Fourier & 35.8 \\
Camera-aware + Linear Projection & 36.0 \\
\midrule
\end{tabular}
\caption{Ablations of the camera-aware positional embeddings. The first row compares to a model that does not use camera-aware positional embedding thus only uses image features as attention keys.
The second row uses a learned embedding for each camera.
The third row uses a camera-aware positional embedding with a random Fourier projection.
The last row uses a camera-aware positional embedding with a linear projection (default).}
\label{table:positional}
\end{table}

\subsection{Camera-aware positional embeddings}

As we have previously seen, the camera-aware positional embedding plays a major role in the success of the cross-view transformer.
Table~\ref{table:positional} compares different choices for this embedding.
We ablate just the positional embedding and keep all other model and training parameters fixed.

Not using any positional embedding performs poorly.
The attention mechanism has a hard time localizing features and identifying cameras.
A learned embedding per camera performs surprisingly well.
This is likely because the camera calibration stays mostly static and a learned embedding simply bakes in all geometric information.
A camera-aware embedding with either a linear or Random Fourier projection~\cite{tancik2020fourier} performs best.
This should not come as a surprise as both can learn a compact embedding that directly captures the geometry of the scene.

\subsection{Accuracy vs distance}

\begin{figure}[t]
\centering
\includegraphics[width=0.75\linewidth]{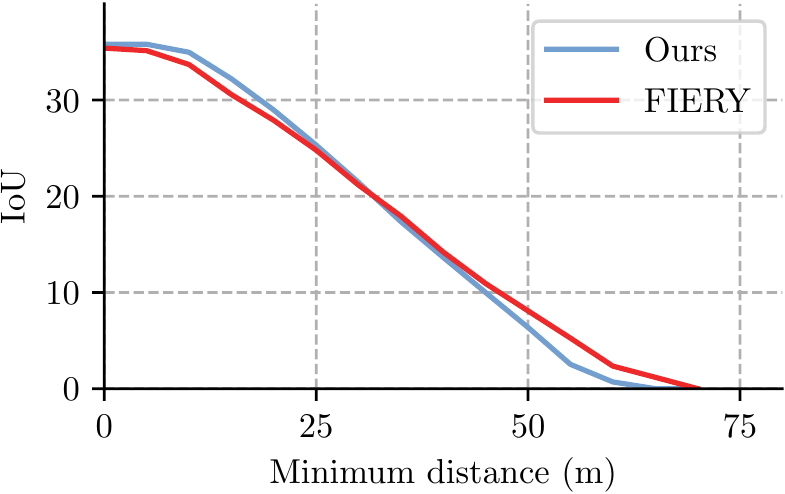}
\caption{A comparison of model performance vs distance to the camera. Each entry shows the average intersection over union accuracy for annotations that are at least distance $d$ away.
}
\label{fig:distance}
\end{figure}

\begin{figure}[b]
\centering
\begin{tikzpicture}[trim left=0.00em]
\begin{axis}[
    small,
    xlabel near ticks,
    ylabel near ticks,
    xlabel=\# Cameras Dropped,
    ylabel=IoU,
    xmin=0, xmax=3,
    ymin=15, ymax=37.5,
    xtick={0, 1, 2, 3},
    xticklabels={0, 1, 2, 3},
    ytick={20, 25, 30, 35},
    width=0.7\linewidth]
\addplot[smooth,mark=*,blue] plot coordinates {
    (0, 36.0)
    (1, 31.2)
    (2, 27.6)
    (3, 22.5)
};
\end{axis}
\end{tikzpicture}
\caption{Degradation of our model as we randomly drop out $m \in \{0, 1, 2, 3\}$ cameras.
The models performance drops linearly as the observed area shrinks roughly linearly with the number of cameras removed.}
\label{fig:cam_drop}
\end{figure}
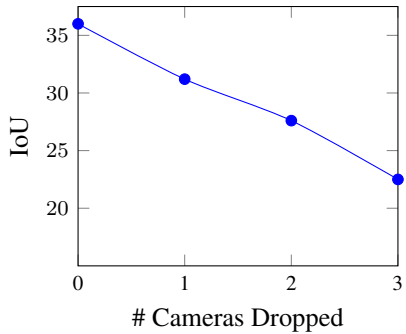

\begin{figure*}[!t]
\centering

\frame{\includegraphics[height=3.3cm, keepaspectratio]{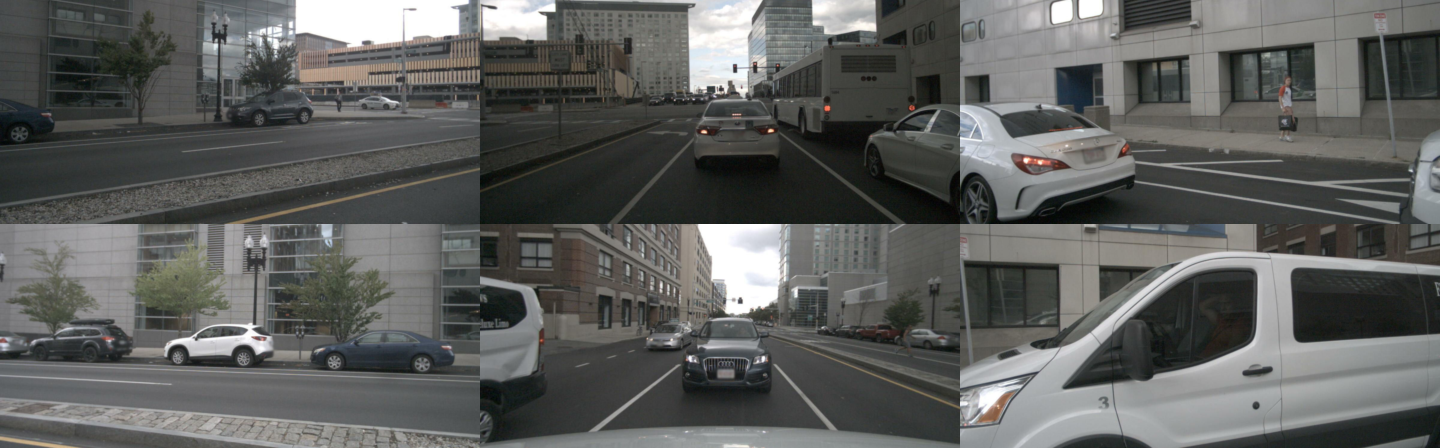}}
\hfill \frame{\includegraphics[height=3.3cm, keepaspectratio]{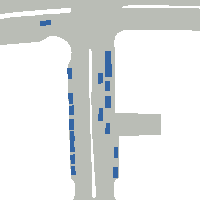}}
\hfill \frame{\includegraphics[height=3.3cm, keepaspectratio]{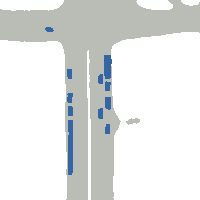}}
\vspace{0.06cm}

\frame{\includegraphics[height=3.3cm, keepaspectratio]{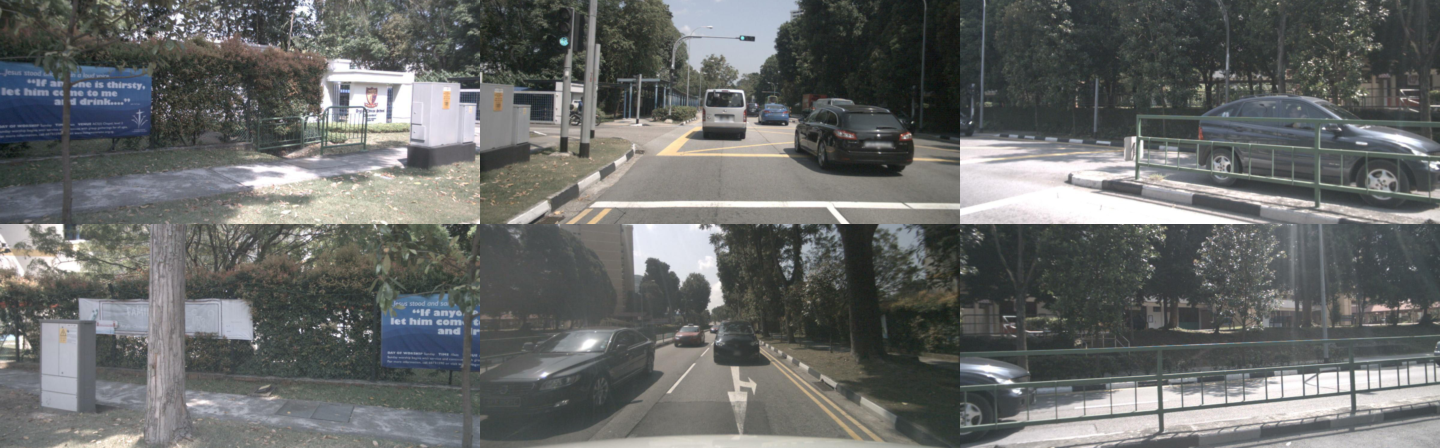}}
\hfill \frame{\includegraphics[height=3.3cm, keepaspectratio]{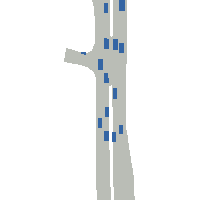}}
\hfill \frame{\includegraphics[height=3.3cm, keepaspectratio]{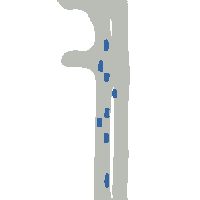}}
\vspace{0.06cm}

\frame{\includegraphics[height=3.3cm, keepaspectratio]{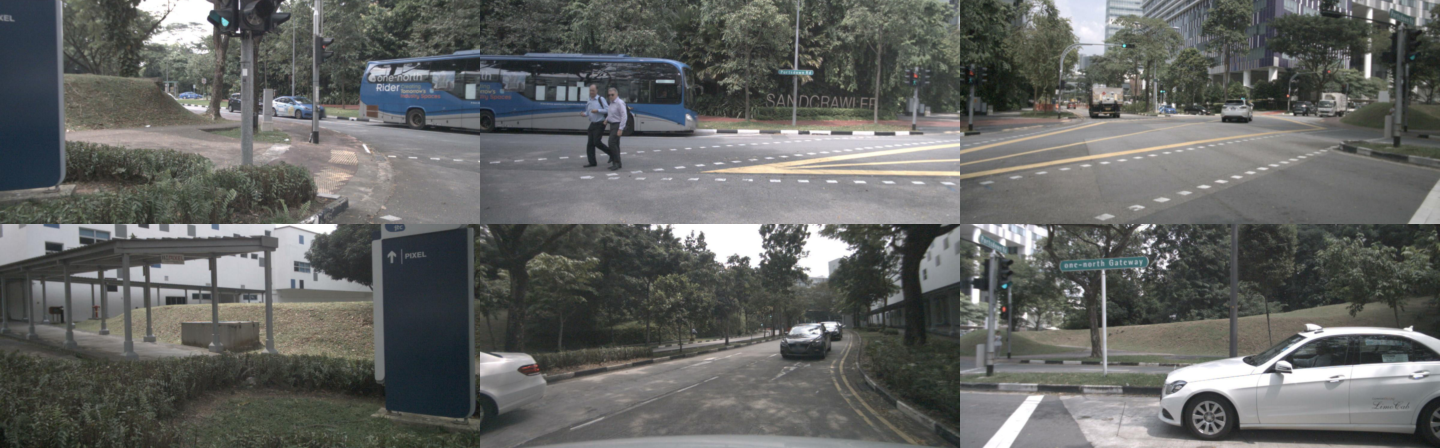}}
\hfill \frame{\includegraphics[height=3.3cm, keepaspectratio]{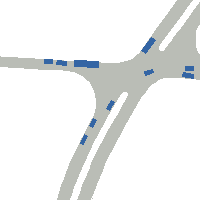}}
\hfill \frame{\includegraphics[height=3.3cm, keepaspectratio]{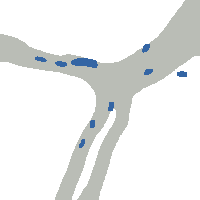}}
\vspace{0.06cm}

\frame{\includegraphics[height=3.3cm, keepaspectratio]{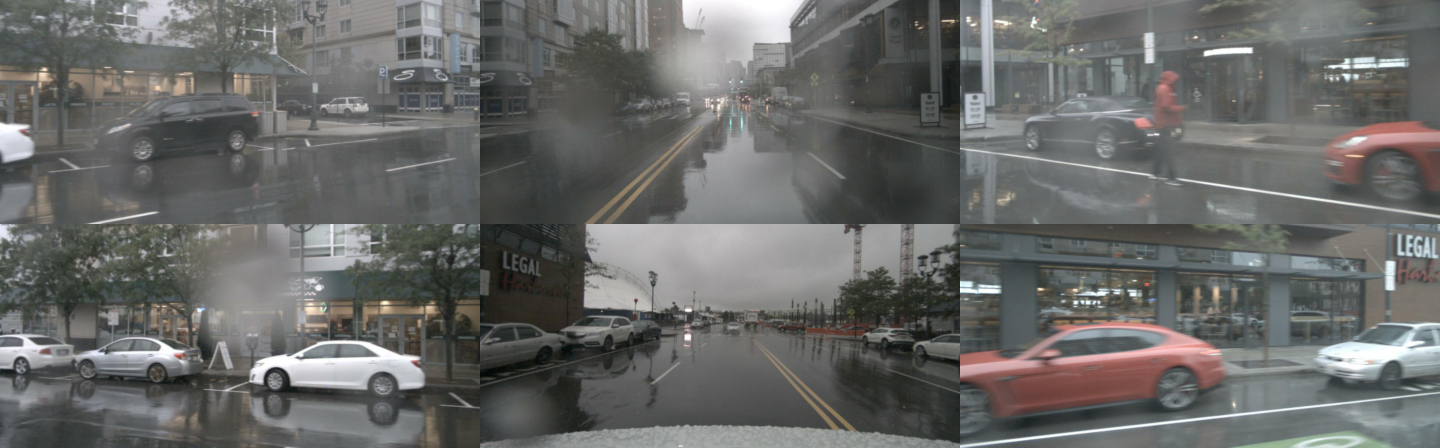}}
\hfill \frame{\includegraphics[height=3.3cm, keepaspectratio]{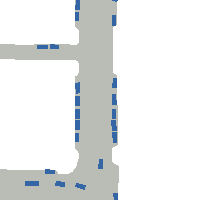}}
\hfill \frame{\includegraphics[height=3.3cm, keepaspectratio]{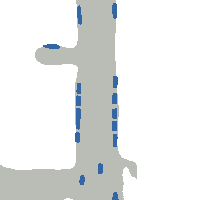}}
\vspace{0.06cm}

\frame{\includegraphics[height=3.3cm, keepaspectratio]{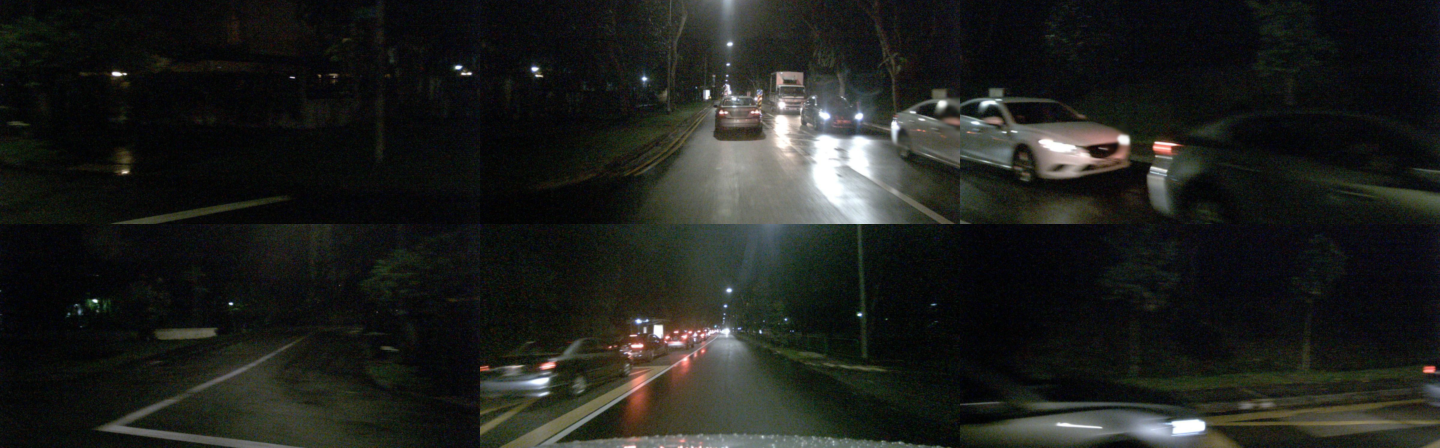}}
\hfill \frame{\includegraphics[height=3.3cm, keepaspectratio]{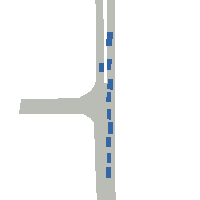}}
\hfill \frame{\includegraphics[height=3.3cm, keepaspectratio]{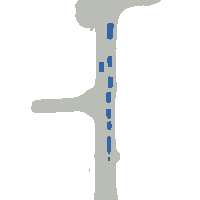}}
\hfill
\vspace{0.06cm}

\caption{Qualitative results on scenes with varying degrees of occlusion. Left shows the six camera views surrounding the vehicle. The top 3 views are front-facing, the bottom 3 views back-facing. On the right is our predicted map-view segmentation for vehicles and driveable area. Second from the right is the ground truth segmentation for reference.
The ego-vehicle is located at the center of the map.}
\label{fig:qualitative}
\vspace{-0.5em}
\end{figure*}

\begin{figure*}[!t]
\centering
\includegraphics[page=1,width=0.85\linewidth]{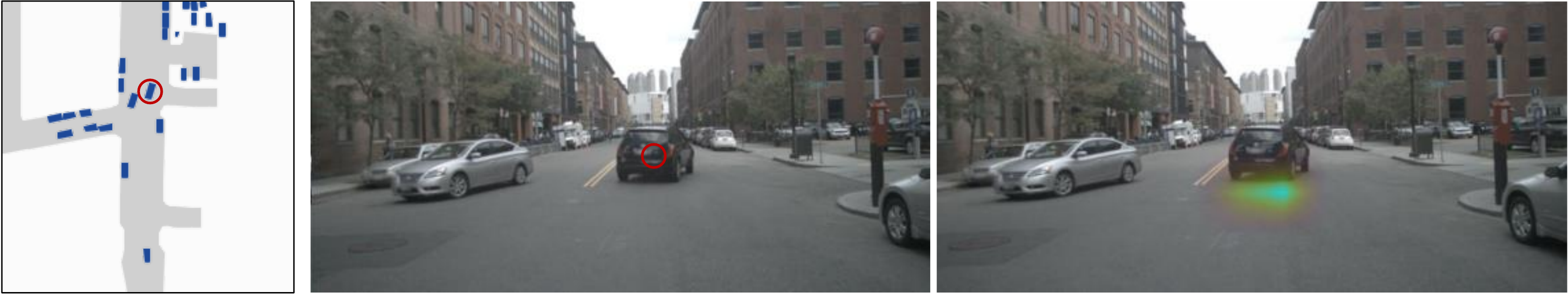}
\includegraphics[page=2,width=0.85\linewidth]{figures/attention}
\includegraphics[page=5,width=0.85\linewidth]{figures/attention}
\includegraphics[page=6,width=0.85\linewidth]{figures/attention}
\caption{Visualization of cross-view attention. We compute attention from a point in map-view coordinates and visualize the corresponding attention values of the front camera view. Note how the network learns geometric correspondences through this attention mechanism.}
\label{fig:geometry}
\end{figure*}

Next, we evaluate how well our model performs as the distance to the ego-vehicle increases.
For this experiment, we measure the intersection-over-union accuracy, but ignore all predictions that are closer than a certain distance to the ego-vehicle.
Figure~\ref{fig:distance} compares to our closest competitor FIERY.

Both models have close to identical error modes.
As the distance to the camera increases, the models get less accurate.
This is easiest explained through actual qualitative results in Figure~\ref{fig:qualitative}.
Farther away vehicles are often (partially) occluded and thus much harder to detect and segment.
Our approach degrades slower for close-by distances, but slightly under-performs FIERY at longer ranges.

Partially occluded far-away samples have fewer corresponding image features, thus learning a mapping from map-view to camera-view directly is harder: There is less training data and fewer geometric priors to rely upon for our model.
We anticipate more data to make up for this difference.

\subsection{Robustness to sensor dropout}

We take a model trained on all six inputs and evaluate the intersection over union (IoU) metric by randomly dropping $m$ cameras for each sample in the validation set.
Figure~\ref{fig:cam_drop} shows how the performance decreases linearly with the number of cameras dropped.
This is quite intuitive as different cameras only overlap marginally.
Thus each removed camera reduces the visible area linearly and drops the performance in the unobserved area.
Note that the transformer-based model is generally quite robust to this camera dropout and the overall performance does not degrade beyond unobserved parts of the scene.

\subsection{Qualitative Results}

Figure~\ref{fig:qualitative} shows qualitative results on a variety of scenes.
For each row, we show the six input camera views and the predicted map-view segmentation along with the ground truth segmentation.
Our presented method can accurately segments nearby vehicles, but does not sense far away or occluded vehicles well.

\subsection{Geometric reasoning in cross-view attention}

Our quantitative experiments indicate that cross-view attention can learn some geometric reasoning.
In Figure~\ref{fig:geometry}, we visualize the image-view attention for several points in the map-view.
Each point corresponds to part of a vehicle.
From qualitative evidence, the attention mechanism can highlight closely corresponding map-view and camera-view locations.

\section{Conclusion}

We present a novel map-view segmentation approach based on a cross-view transformer architecture built on top of camera-aware positional embeddings.
The proposed approach achieves state-of-the-art performance, is simple to implement, and runs in real time.

\paragraph{Acknowledgements.} This material is based upon work supported by the National Science Foundation under Grant No. IIS-1845485 and IIS-2006820.

\pagebreak

{\small
\bibliographystyle{ieee_fullname}
\bibliography{egbib}
}

\end{document}